\documentclass{article}

\usepackage{arxiv}

\usepackage[utf8]{inputenc} % allow utf-8 input
\usepackage[T1]{fontenc}    % use 8-bit T1 fonts
\usepackage{hyperref}       % hyperlinks
\usepackage{url}            % simple URL typesetting
\usepackage{booktabs}       % professional-quality tables
\usepackage{amsmath,amssymb,amsfonts}       % blackboard math symbols
\usepackage{nicefrac}       % compact symbols for 1/2, etc.
\usepackage{microtype}      % microtypography

\usepackage{algorithmic}
\usepackage{subcaption}     % subfigure
\usepackage{booktabs}       % booktabs table
\usepackage{multirow}       % multirow in tables
\usepackage{verbatim}       % multiline comment
\usepackage{makecell}       % multiline cells
\usepackage{cite}
\usepackage{nameref,hyperref}
\usepackage[table]{xcolor}
\usepackage{array}
\usepackage{float}
\usepackage[aboveskip=1pt,labelfont=bf,labelsep=period,justification=raggedright,singlelinecheck=off]{caption}
\usepackage{lastpage,fancyhdr,graphicx}
\graphicspath{{figures/}}

\title{Principal component-based image segmentation: a new approach to outline in vitro cell colonies}

\author{
  Delmon Arous\thanks{The author is further affiliated with the Department of Medical Physics, Oslo University Hospital, Oslo} \\
  Department of Physics\\
  University of Oslo\\
  Oslo, Norway \\
  \texttt{delmon.arous@fys.uio.no} \\
   \And
  Stefan Schrunner\thanks{The author is currently affiliated with the Department of Data Science, Norwegian University of Life Sciences, Ås, Norway} \\
  Department of Medical Physics\\
  Oslo University Hospital\\
  Oslo, Norway\\
  \texttt{stefan.schrunner@nmbu.no} \\
   \AND
  Ingunn Hanson \\
  Department of Physics\\
  University of Oslo\\
  Oslo, Norway \\
   \texttt{ingunn.hanson@fys.uio.no} \\
   \And
  Nina F.J. Edin \\
  Department of Physics\\
  University of Oslo\\
  Oslo, Norway \\
   \texttt{n.f.j.edin@fys.uio.no} \\
   \And
  Eirik Malinen\footnotemark[1] \\
  Department of Physics\\
  University of Oslo\\
  Oslo, Norway \\
   \texttt{eirik.malinen@fys.uio.no} \\
}

\begin{document}
\maketitle

\begin{abstract}
The \textit{in vitro} clonogenic assay is a technique to study the ability of a cell to form a colony in a culture dish. By optical imaging, dishes with stained colonies can be scanned and assessed digitally. Identification, segmentation and counting of stained colonies play a vital part in high-throughput screening and quantitative assessment of biological assays. Image processing of such pictured/scanned assays can be affected by image/scan acquisition artifacts like background noise and spatially varying illumination, and contaminants in the suspension medium. Although existing approaches tackle these issues, the segmentation quality requires further improvement, particularly on noisy and low contrast images. In this work, we present an objective and versatile machine learning procedure to amend these issues by characterizing, extracting and segmenting inquired colonies using principal component analysis, $k$-means clustering and a modified watershed segmentation algorithm. The intention is to automatically identify visible colonies through spatial texture assessment and accordingly discriminate them from background in preparation for successive segmentation. The proposed segmentation algorithm yielded a similar quality as manual counting by human observers. High $F_1$ scores ($>0.9$) and low root-mean-square errors (around $14\%$) underlined good agreement with ground truth data. Moreover, it outperformed a recent state-of-the-art method. The methodology will be an important tool in future cancer research applications.
\end{abstract}

% keywords can be removed

\section{Introduction}

Clonogenic assay or colony formation assay serves as a means to assess viable, growing cell colonies \cite{franken2006clonogenic} and plays imperative roles in radiobiology \cite{moiseenko2007vitro}, microbiology \cite{krastev2011systematic} and immunology \cite{junkin2014microfluidic}. Manual identification of colonies (conglomerations composed of $>50$ cells) is time-consuming with potentially large inter-observer variations. High-pass optical image scanners, digital cameras or other imaging systems introduces a new field of image processing solutions. However, digital assessment of inspected colonies depends on several factors such as background noise, clustering of cells/colonies, variable colony confluency and colony specific features including size and circularity. Therefore, it is essential to have a robust and adaptive approach that takes these discernments into consideration and that provides accurate, fast, objective and reliable segmentation of colonies. 

Automated cellular and bacterial colony counters have been an abiding issue of interest \cite{mansberg1957automatic}. There are currently commercial solutions available, but these are proprietary tools that require purchase of respective imaging stations and may be cost-prohibitive. In addition, these products are running segmentation algorithms that are undisclosed, making them restrictive and hard to interpret for the user.

Contrarily, several free and open-source colony segmentation methods are accessible for the user as they are supported on common operating systems. Applications within this category includes Circular Hough image Transform Algorithm (CHiTA) \cite{bewes2008automated} and NIST’s Integrated Colony Enumerator (NICE) \cite{clarke2010low}. CHiTA identifies cell colonies by intensity gradient field discrimination. However, the utilization of the circular Hough transform makes the program prone to neglect more elongated segments. NICE operates by combining extended-minima transform and thresholding algorithms. The extended-minima analysis is used to find the center of the colonies and to distinguish adjacent colonies. Nonetheless, this segmentation approach does not take different colony shapes, sizes or variable staining into account, which could render the following intensity threshold faulty. 

OpenCFU is a cross-platform and C++ based open-source software, made freely available \cite{geissmann2013opencfu}. It declares to be faster, more accurate and more robust to the presence of artifacts compared to NICE. The application is operated via an intuitive graphical user interface (GUI) which is also extensively described in a user manual. Although the program is able to initiate a batch acquisition and exclude morphologically anomalous objects, the selection method is restricted to circular objects. This could be a concern when processing cell lines with non-circular colony phenotype.

CellProfiler is another free, open-source program that addresses a variety of biological features, including standard and complex morphological assays (e.g., cell count, size, cell/organelle shape, protein staining) \cite{carpenter2006cellprofiler}. The program uses either standardized pipelines or individual modules that can be customized to specific tasks. However, due to the sequential order of the modules, the performance of the cumulative operations may not be optimal. Furthermore, Deep Convolutional Neural Networks (DCNNs) have been combined with common automated pipelines in CellProfiler to solve segmentation tasks. Still, the DCNN performance is strongly dependent on the availability of large amounts of high-quality and problem-specific training data \cite{sadanandan2017automated}.

More recently, the state-of-the-art method, AutoCellSeg, was developed utilizing adaptive multi-thresholding to extract connected cell colony conglomerations of interest and automatic feedback-based watershed segmentation to further partition the conglomerations into separate colonies \cite{torelli2018autocellseg}. This algorithm was applied on images of four different types of bacterial species, where the results were tested against established ground truths (GTs) showing greater accuracy performance than OpenCFU and CellProfiler.

We propose a versatile automated segmentation method with an image analysis pipeline consisting of signal decomposition of the raw input image, foreground-background separation, segmentation of the colonies, feature extraction and post-segmentation correction.
In essence, the segmentation procedure relies on three key techniques:
\begin{enumerate}
    \item \textbf{Principal component analysis (PCA)} - by image channel decomposition to convert information stored in the color channels into different contrast planes, whereby automated channel selection is performed by spatial texture analysis using the gray-level co-occurence matrix (GLCM),
    \item $\mathbf{k}$\textbf{-means clustering} - for vector quantization of computed PCA channel intensity pixels to mask out connected colonies from background,
    \item \textbf{multi-threshold-based watershed segmentation} - to further segment the extracted features into colonies by incorporating \textit{fuzzy logic}. 
\end{enumerate}

\begin{figure}[ht!]
    \centering
    \includegraphics[width=1.0\textwidth]{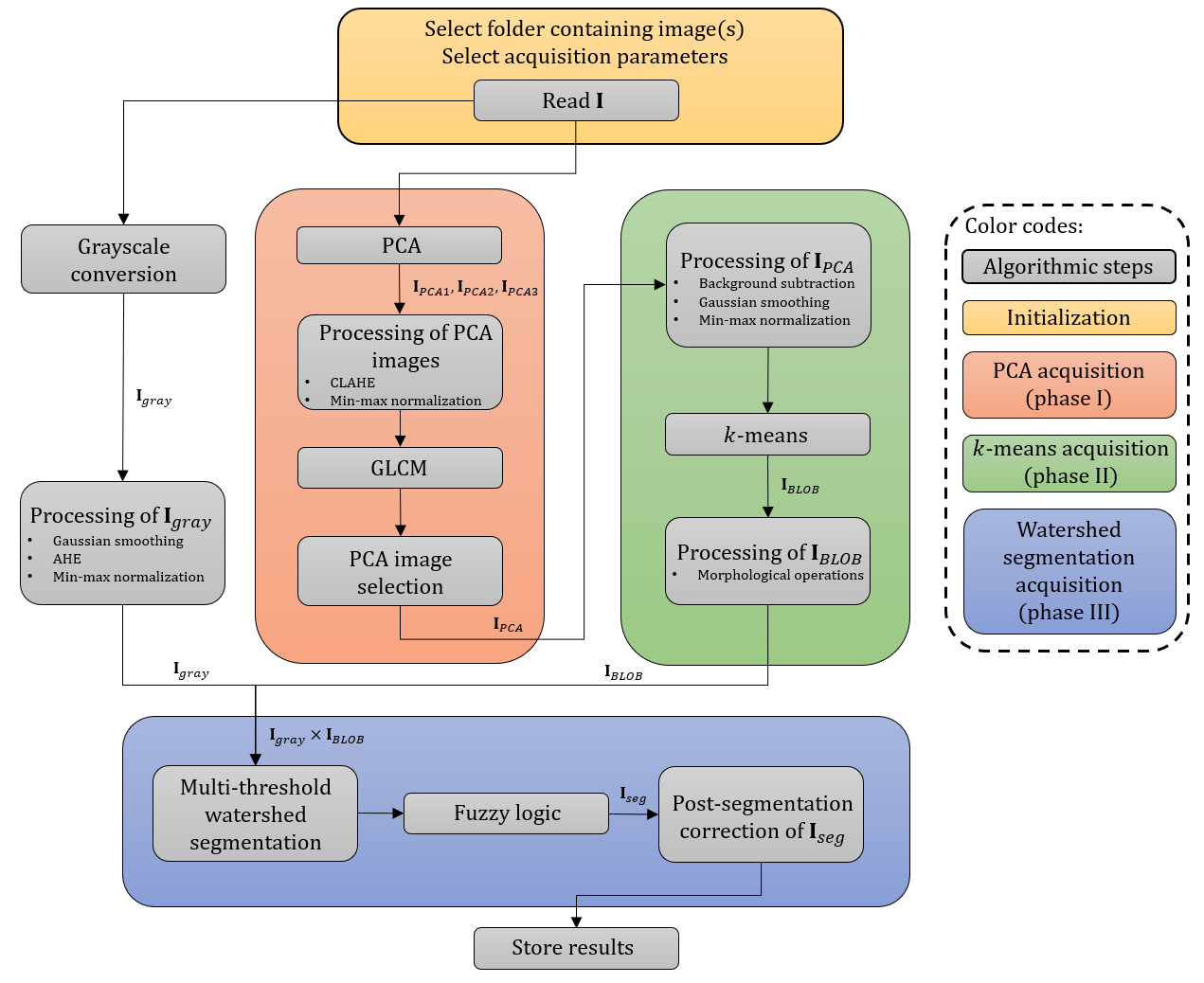}
    \caption{Overview of the image processing pipeline showing the main steps. Initially, the $rgb$ image, $\mathbf{I}$, is read from the selected folder, where segmentation parameters are also chosen by the user in the initialization. Phase I: the color components of $\mathbf{I}$ are decomposed before performing a principal component analysis (PCA) on the input data. The principal component (PC) images, $\mathbf{I}_{PCA1}$, $\mathbf{I}_{PCA2}$ and $\mathbf{I}_{PCA3}$, of the $rgb$ input sample are then processed, by means of contrast-limited adaptive histogram equalization (CLAHE), prior to texture analysis via gray-level co-occurrence matrix (GLCM) computation. Phase II: from the GLCM-analysis, the channel with minimum contrast is selected, $\mathbf{I}_{PCA}$, and supplied to the $k$-means analysis phase. The raw PC image is processed in order to augment the foreground information from the background, whilst restraining background information. Performing $k$-means yields a binary image of the merged colonies, $\mathbf{I}_{BLOB}$. Phase III: multiplying $\mathbf{I}_{BLOB}$ by the grayscale image of $\mathbf{I}$, $\mathbf{I}_{gray}$, masks out the relevant intensity regions in preparation for watershed segmentation. Multiple intensity-thresholds are imposed on each inquired region, where respective colony features are evaluated using fuzzy logic providing a segmented binary image of $\mathbf{I}_{BLOB}$, $\mathbf{I}_{seg}$. Finally, $\mathbf{I}_{seg}$ is corrected post-segmentation before the conclusive results (colony count, features etc.) are saved as \textit{.csv} files.}
    \label{fig: pipeline}
\end{figure}

In the present study, we show the applicability of each separate method as to supply linked information downstream of the image analysis pipeline (see \autoref{fig: pipeline}). Hence, a collective integration of these techniques to assess the colony viability yields a novel approach that is presently evaluated. Specifically, the inherent information provided by the principal component (PC) channel that serves as an explicit depiction of the colonies is automatically selected by computing the GLCM. This selection is used as a basis for the watershed segmentation procedure, which has not been addressed previously. Subsequent segmentation optimization takes into account cell colony characteristics such as e.g. circularity and size through adaptive fuzzy logic consensus for each individual image. By forming a fuzzy mathematical description of the selection space for each feature, aggregate colony feature scores are computed to objectively choose the optimal watershed segmentation outcome. The performance of this approach is evaluated against a state-of-the-art methodology, as well as manual cell colony count on a selection of datasets showing different characteristics.

With the presented methodology, we circumvent drawbacks of the discussed algorithms such as disregard of geometrical shape, basic one-dimensional thresholding, convoluted parameter settings and the necessity of high amounts of training data. As will become evident, our colony segmentation method - the automated colony counting (ACC) algorithm - accurately maps cell colonies and yields quantitative estimates of number, localization and density. Moreover, since the AutoCellSeg method was reported to outperform other methodologies, our current ACC procedure was chosen to be benchmarked against this approach.

\section{Methods}

\subsection{Image channel decomposition}

We apply a decomposition method to the multivariate data composed of the $p=3$ color channels. The idea is to identify the information about cell colonies and separate it from cell flask, shadows and noise. Originally, all of these signals are distributed across the 3 channels of the truecolor image resulting from an optical scan of a cell flask containing stained colonies (see subsection \ref{data description}). The proposed algorithm de-mixes the signal via a linear combination of sources using PCA. With this approach, we map colony information on a single plane by bundling the information from all color channels \cite{Lay2020}.

Let $\mathbf{X}_i$ denote the observation vector in $\mathbb{R}^p$ comprising the red ($r$), green ($g$) and blue ($b$) color components of the $i$th pixel in the $M \times N$ input image, $\mathbf{I}$. By rearranging the multichannel components,
the matrix of observations, $\mathbf{X} \in \mathbb{R}^{p\times MN}$, is then defined to be a matrix of the form
\begin{equation} \label{eq: X_RGB}
\mathbf{X} = 
\left[ \begin{matrix} 
\mathbf{X}_1 & \mathbf{X}_2 & \cdots & \mathbf{X}_{MN} \end{matrix} \right] 
= \left[ \begin{matrix} 
r_1 & r_2 & \cdots & r_{MN} \\ 
g_1 & g_2 & \cdots & g_{MN} \\ 
b_1 & b_2 & \cdots & b_{MN} 
\end{matrix} \right].
\end{equation}
The mean-deviation form matrix $\hat{\mathbf{X}} \in \mathbb{R}^{p \times MN}$ of $\mathbf{X}$ is introduced as ${\hat{\mathbf{X}}}_i = \mathbf{X}_i - \boldsymbol{\mu}$, for $i=1,\ldots,MN$, where $\boldsymbol{\mu}$ is the sample mean of the observation matrix $\mathbf{X}$. Consequently, $\hat{\mathbf{X}} \in \mathbb{R}^{p \times MN}$ is introduced as 
\begin{equation} \label{eq: Xhat_MeanDeviationForm}
\hat{\mathbf{X}} = \left[ \begin{matrix} {\hat{\mathbf{X}}}_1 & {\hat{\mathbf{X}}}_2 & \cdots & {\hat{\mathbf{X}}}_{MN} \end{matrix}\right].
\end{equation}

\subsubsection{Principal component analysis (PCA)}
PCA is a popular method for extracting relevant information from multivariate data, mainly focusing on dimensionality reduction \cite{wold1987principal, abdi2010principal}. It aims to transform input variables linearly into PCs, sorted by their explained variance in a descending order. The main idea is that a high percentage of the total variance of the input data is covered by the first output PCs.

Technically, PCA describes the change of variable for each observation vector of $\hat{\mathbf{X}}$ by,
\begin{equation} \label{eq: VariableChange}
{\hat{\mathbf{X}}}_i = \left[\begin{matrix} 
{\hat{x}}_{i1} \\ 
{\hat{x}}_{i2} \\ 
\vdots \\ 
{\hat{x}}_{ip} \end{matrix} \right] = 
\left[ \begin{matrix} 
\mathbf{u}_1 & \mathbf{u}_2 & \cdots & \mathbf{u}_p \end{matrix} \right] \left[ \begin{matrix} 
{\hat{y}}_{i1} \\
{\hat{y}}_{i2} \\
\vdots \\
{\hat{y}}_{ip} \end{matrix} \right] = 
\mathbf{P}{\hat{\mathbf{Y}}}_i,
\end{equation}
where the orthogonal matrix $\mathbf{P} =\left[\begin{matrix}\mathbf{u}_1&\cdots&\mathbf{u}_p\\\end{matrix}\right] \in \mathbb{R}^{p \times p} $ consists of the unit eigenvectors (or PCs) of the co-variance matrix of $\hat{\mathbf{X}}$, $\mathbf{C} \in \mathbb{R}^{p \times p}$, determined via singular value decomposition (SVD) of $\mathbf{C}$. Since $\mathbf{P}$ is an invertible matrix, a linear combination of the original variables in ${\hat{\mathbf{X}}}_i$ determines the new PC pixel values – the intensity variation of each composite $rgb$ pixel – by the variable transformation,
\begin{align}
{\hat{y}}_{i1} 
&= \mathbf{u}_1^T {\hat{\mathbf{X}}}_i 
= u_1^{(1)}{\hat{x}}_{i1} + u_2^{(1)}{\hat{x}}_{i2} + \cdots + u_p^{(1)}{\hat{x}}_{ip}, \label{eq: PCA1} \\
{\hat{y}}_{i2} 
&= \mathbf{u}_2^T {\hat{\mathbf{X}}}_i 
= u_1^{(2)}{\hat{x}}_{i1} + u_2^{(2)}{\hat{x}}_{i2} + \cdots + u_p^{(2)}{\hat{x}}_{ip}, \label{eq: PCA2} \\
{\hat{y}}_{i3} 
&= \mathbf{u}_3^T {\hat{\mathbf{X}}}_i 
= u_1^{(3)}{\hat{x}}_{i1} + u_2^{(3)}{\hat{x}}_{i2} + \cdots + u_p^{(3)}{\hat{x}}_{ip}, \label{eq: PCA3}
\end{align}
where $u_1^{(1)},\ldots,u_p^{(1)}$, $u_1^{(2)},\ldots,u_p^{(2)}$ and $u_1^{(3)},\ldots,u_p^{(3)}$ are the entries in the 1st, 2nd and 3rd PC vector, $\mathbf{u}_1$, $\mathbf{u}_2$ and $\mathbf{u}_3$ respectively, while the new variables ${\hat{y}}_{i1}$, ${\hat{y}}_{i2}$ and ${\hat{y}}_{i3}$ represent the 1st, 2nd and 3rd PC pixel values given by ${\hat{\mathbf{Y}}}_i = \mathbf{P}^T {\hat{\mathbf{X}}}_i$ from equation \eqref{eq: VariableChange}. This projects an image in the 1st, 2nd and 3rd dimension of the PCA space - $\mathbf{I}_{PCA1}$, $\mathbf{I}_{PCA2}$ and $\mathbf{I}_{PCA3}$ respectively - reflecting the triplet color variation of the inquired image.

\subsubsection{Gray-level co-occurrence matrix (GLCM)}

In our application, the PC images ($\mathbf{I}_{PCA1}$, $\mathbf{I}_{PCA2}$, $\mathbf{I}_{PCA3}$) include variance information about the cell colonies, cell container, shadows and noise. Among the PC images, we assume that only one of the images offers a reliable and selective depiction of the colonies, whereas the two remaining PC images contain (variance) information representing other image contributions.

The GLCM is a statistical approach for analyzing texture \cite{haralick1973textural, haralick1992computer}. We will use image contrast, as defined from the GLCM, to identify the optimal PC image with respect to cell colony depiction. In a single input channel image (representing in our case one PC image), $\mathbf{J}$, the co-occurrence matrix, $\mathbf{G} \in \mathbb{R}^{N_g \times N_g}$, is defined as the frequency of pixel-pairs along a particular distance and direction in $\mathbf{J}$ of $N_g$ gray-levels:
\begin{align}
g_{ij}\left(d,\theta \right)
&= \sum_{x=1}^{N} \sum_{y=1}^{M} \left\{
    \begin{array}{ll}
      1, & \text{if } J(x,y)=i \text{ and } J(x+d \cos{\theta},y+ d \sin{\theta})=j, \\
      0, & \text{otherwise,} 
   \end{array}
   \right. \label{eq: GLCM} \\
\Tilde{g}_{ij}\left(d,\theta \right)
&= \frac{g_{ij}\left(d,\theta \right)}{\sum_{i=1}^{N_g} \sum_{j=1}^{N_g} g_{ij}\left(d,\theta \right) } , \label{eq: NGLCM}
\end{align}
where $g_{ij}\left(d,\theta \right)$ and $\Tilde{g}_{ij}\left(d,\theta \right)$ denotes the $(i,j)$-th entry in the co-occurrence matrix and normalized co-occurrence matrix, respectively. The GLCM describes the relative frequency between the pixel-pair $(x,y)$ and $(x+d \cos{\theta}, y+ d \sin{\theta})$ separated by a specified displacement $d$ and angle $\theta$ - offset - with gray-level intensity $i$ and $j$, respectively, in the domain ${i,j} \in {1,2,...,N_g}$. 

Next, the Haralick feature \cite{haralick1973textural} for contrast is computed from the GLCM as a statistical measure to describe colony texture characteristic and is used for PC selection: 
\begin{equation} \label{eq: Contrast}
\text{Contrast}_{\mathbf{J}} = \sum_{i=1}^{N_g} \sum_{j=1}^{N_g} \left| i-j \right|^2 \Tilde{g}_{ij}(d,\theta).
\end{equation}
It returns a measure of the intensity contrast repetition rate for a pixel-pair across the whole image. This statistic ranges in the interval $\left[0, (N_g-1)^2 \right]$, where it is 0 for a constant image. Therefore, low contrast entails an image that features low spatial frequencies.

The PC selection criterion involves choosing the PC image with the lowest contrast statistic. As either $\mathbf{I}_{PCA1}$, $\mathbf{I}_{PCA2}$ or $\mathbf{I}_{PCA3}$ expresses the color variation of solely the colonies, the most suitable PC image is composed of pixel values that are insensitive to and suppress the presence of various high-contrast artifacts such as contaminants/residue in the suspension medium, inevitable shadow artifacts due to imaging/scanning procedures, inherent background noise emanated from the image/scan acquisition and the cell container boundary. Hence, the spatial frequency of local color variations depicting merely the colonies is minimum in the PC image characterizing the colonies relative to the remaining two PCs depicting all other elements. Hence, the PC channel with the lowest contrast estimation through equation \eqref{eq: Contrast} results in the PC image selection describing the colonies optimally:
\begin{equation} \label{eq: Selection}
\mathbf{I}_{PCA} 
= \underset{X \in \{ \mathbf{I}_{PCA1}, \mathbf{I}_{PCA2}, \mathbf{I}_{PCA3} \}}{\arg\min} \text{Contrast}_{X}.
\end{equation}

Prior to GLCM contrast estimation, each PC image is enhanced by applying contrast-limited adaptive histogram equalization (CLAHE) \cite{zuiderveld1994contrast} to aid the selection criterion in equation \eqref{eq: Selection}. Through dividing an image into a grid of rectangular regions, the histogram of the contained pixels for each region is computed. The contrast of each region is locally optimized by redistributing the pixel intensity according to a transform function, where a uniform histogram equalization distribution is used here. Then, by imposing a clip limit (or contrast factor) as a maximum on the computed histograms, over-saturation of particularly homogeneous areas (characterized by high peaks in the contextual histograms) is reduced, which prevents over-enhancement of, e.g., noise and edge-shadowing effect derived from an unlimited adaptive histogram equalization (AHE).

\subsection{k-means clustering}

To distinguish the conglomerate cell colonies characterized in $\mathbf{I}_{PCA}$ from background, we deploy $k$-means clustering \cite{lloyd1982least} on the raw $\mathbf{I}_{PCA}$ to produce a binary mask of the cell colonies. After subtracting the background through opening-closing by reconstruction in order to augment foreground recognition and min-max normalization of the values in $\mathbf{I}_{PCA}$, we construct a feature matrix $\mathbf{Z}$ by aggregating the $i$th pixel value, $p_i$, with its 8-connected neighbors, $p_i^{(1)},\ldots,p_i^{(8)}$. We obtain a $9 \times MN$ matrix,
\begin{equation} \label{eq: Z}
\mathbf{Z} = \left[ \begin{matrix} 
\mathbf{Z}_1 & \mathbf{Z}_2 & \cdots & \mathbf{Z}_{MN} \end{matrix} \right] =
\left[ \begin{matrix} 
p_1         & p_2       & \cdots & p_{MN} \\
p_1^{(1)}   & p_2^{(1)} & \cdots & p_{MN}^{(1)} \\
\vdots      & \vdots    & \ddots & \vdots \\
p_1^{(8)}   & p_2^{(8)} & \cdots & p_{MN}^{(8)}
\end{matrix} \right],
\end{equation}
where each pixel cluster $\mathbf{Z}_i$, $i=1,\ldots,MN$, is assigned to either background, $\mathbf{c}_0=\left[0,\ldots,0\right]^T$, or foreground, $\mathbf{c}_1=\left[1,\ldots,1\right]^T$, through squared Euclidean distance (ED) minimization
\begin{equation} \label{eq: dist_min}
\bar{\mathbf{c}}_{i} = \underset{{\mathbf{c} \in\{\mathbf{c}_0, \mathbf{c}_1\}}}{\arg\min} \left\| \mathbf{Z}_i-\mathbf{c} \right\|^2,
\end{equation}
where $\bar{\mathbf{c}}_{i}$ denotes to the centroid of the class assigned to pixel $i$. Hence, finding the optimal distance by $k$-means ($k=2$) creates a binary mask, $\mathbf{I}_{BLOB}$, containing contiguous colony components denoted as Binary Large OBjects (BLOBs), $BLOB_1,\ldots,\ldots,BLOB_n$, where $n$ is the total number of BLOBs. The BLOB extraction is therefore made independent of geometrical shape as all sizes and shapes with adequate pixel intensity are masked out by $k$-means.

\subsection{Topological multi-threshold watershed segmentation}

We further apply the watershed algorithm following \cite{khan2016new,torelli2018autocellseg}, which we modify and expand to, among other, handle colony confluency. Here, distance transformation along multi-threshold-based watershed is consolidated with quality criteria to recursively subdivide the BLOBs of interest into distinct colonies through \textit{catchment basin} and \textit{watershed line} formulation \cite{GonzalezWoods2018}. 

\begin{figure}[ht!]
    \centering
    \includegraphics[width=0.8\textwidth]{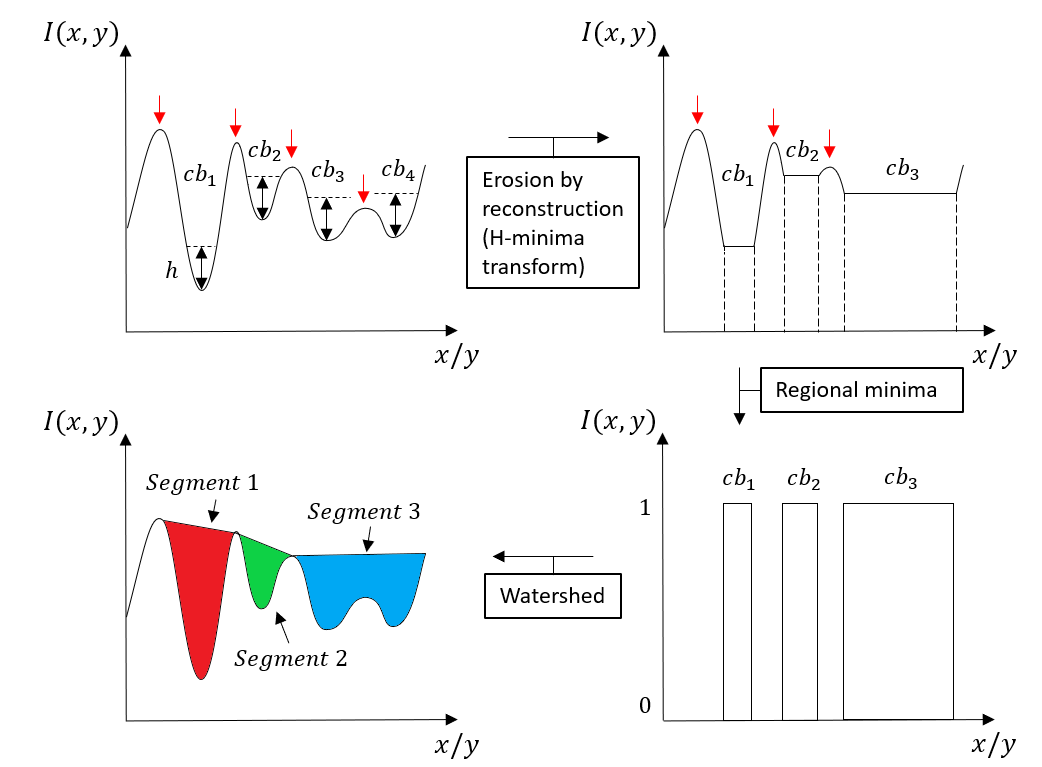}
    \caption{Schematic watershed procedure for a simplified intensity profile along a single dimension. In the top left subfigure, pixels in the intensity distribution are assigned to either a catchment basin ($cb_i$) or a watershed line (red arrows). The height, $h$, signifies the threshold values used for extended-minina transform to morphologically mask out basin regions of interest.}
    \label{fig: watershed}
\end{figure}

The established BLOBs in $\mathbf{I}_{BLOB}$ are divided into individual colonies by the watershed algorithm. Watershed segmentation relies on a topographic (intensity) information across two spatial coordinates, $x$ and $y$, reflecting the colony number in each BLOB. This information is obtained from $\mathbf{I}_{gray}$ since $\mathbf{I}_{PCA}$ is not a measure of colony intensity, but rather a homogeneous variance region of the BLOBs. Thus, by multiplying $\mathbf{I}_{gray}$ with $\mathbf{I}_{BLOB}$, a topographic surface is provided where the background is masked out. However, erroneous over-segmentation may result from direct application of the watershed algorithm due to noise and local irregularities in the intensity distribution. This may accordingly lead to the formation of overwhelming amounts of basin regions. Therefore, we utilize extended-minima transform to avoid the tendency to include regional minima. All regional minima are identified as connected pixels with intensities that differ more than a specified threshold, $h$, relative to neighboring pixels, while the remaining local minima whose depths are too shallow are suppressed (see \autoref{fig: watershed}).

Thus, the definition of the extended-minima operator for a given $h$, $\text{E-MIN}_h$, produces a desired binary mask of the pronounced basins, 
\begin{equation} \label{eq: extended-minima}
\text{E-MIN}_h\left(I(x,y)\right) 
= \text{R-MIN} \left[R_I\left(I\left(x,y\right)+h\right)\right] ,
\end{equation}
where $R_I$ denotes reconstruction by erosion of $I$ from $I+h$ to suppress all shallow minima and $\text{R-MIN}$ represents the regional minima operator of corresponding erosion.

\begin{figure}[H]
    \centering
    \includegraphics[width=1.0\textwidth]{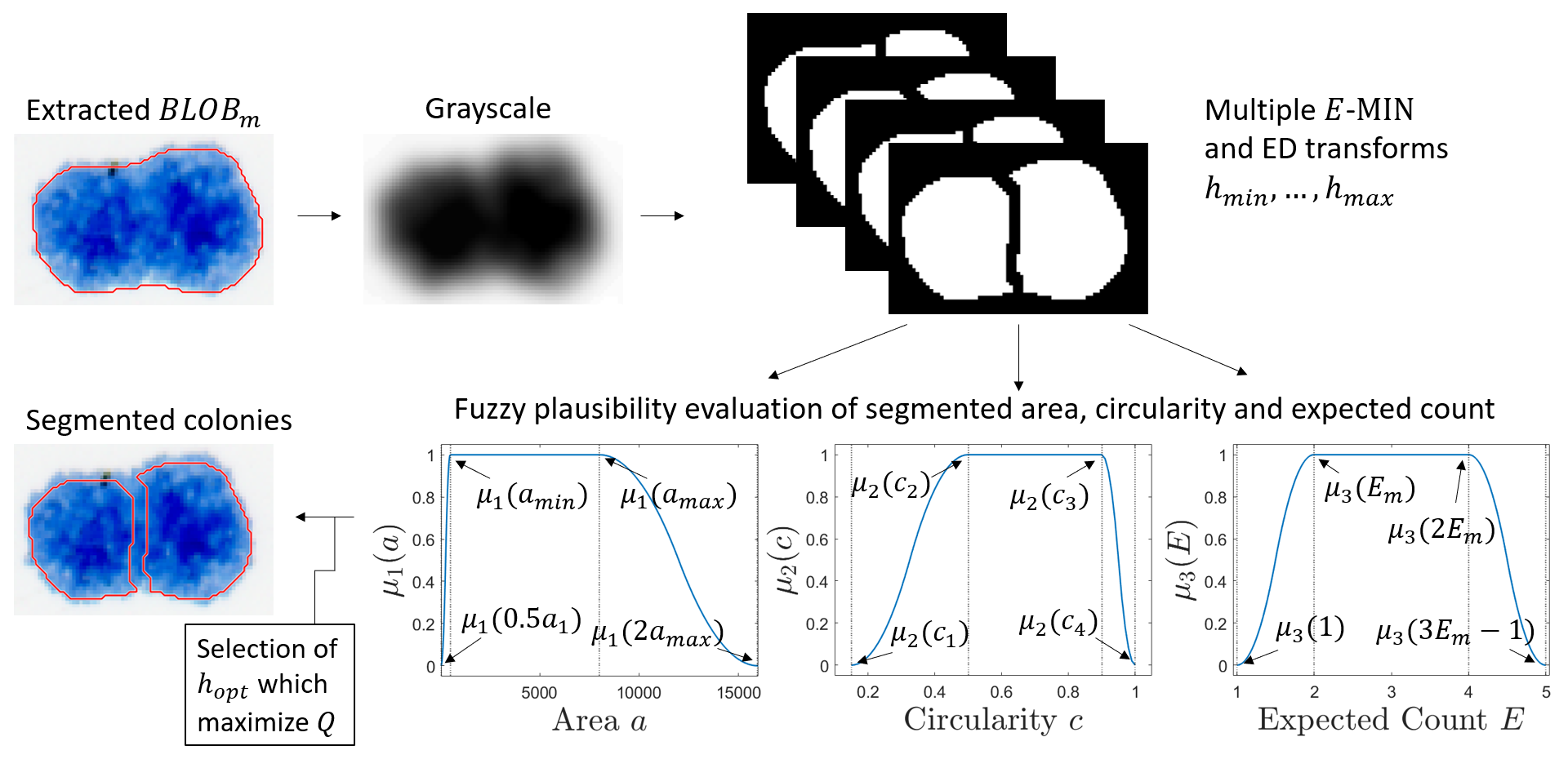}
    \caption{Watershed processing pipeline for a single iterated BLOB, $BLOB_m$. The conglomeration is firstly converted to intensity before applying several extended-minima operators ($\text{E-MIN}$) and Euclidean distance (ED) transforms, where each transformation yields segmented colonies. The validity of each segmentation outcome is subsequently graded using fuzzy pi-shaped MFs $\mu_j \left(u;e_1^{(j)},e_2^{(j)},e_3^{(j)},e_4^{(j)}\right)$ for fuzzy set $j$ representing colony area, circularity and expected count.}
    \label{fig: FuzzyPi}
\end{figure}

Employing $\text{E-MIN}_h$ on $\mathbf{I}_{gray}$ yields varying outcomes for different thresholds, $h$. To account of this, multiple $\text{E-MIN}_{h_i}$, $h_i \in [h_{min},h_{max}]$, are sequentially applied on each $BLOB_m$, for $m=1,\ldots,n$, to create a manifold of candidate segmentation outcomes in form of binary masks. Additionally, to withstand high cell confluency and achieve a proper segmentation, ED transform is conducted on each mask from every $h_i$. Then, the optimal transformation is selected that maximizes the quality segmentation criterion, $Q$, which incorporates fuzzy logic,
\begin{align}
    h_{opt} &= \underset{h_i}{\arg\max} \, Q(h_i) \label{eq: h_opt} \\
    Q &= \mu_1 \cdot \mu_2 \cdot \mu_3 ,\label{eq: Q}
\end{align}
where $\mu_1$, $\mu_2$ and $\mu_3$ are fuzzy spline-based pi-shaped membership functions (MFs) given by
\begin{equation} \label{eq: mu_fuzzy}
    \mu_j\left(u \right) = \left\{
    \begin{array}{lc}
      2\left(\frac{u-e_1^{(j)}}{e_2^{(j)}-e_1^{(j)}}\right)^2, & e_1^{(j)} \leq u \leq \frac{e_1^{(j)} + e_2^{(j)}}{2} \\
      1 - 2\left(\frac{u-e_2^{(j)}}{e_2^{(j)}-e_1^{(j)}}\right)^2, & \frac{e_1^{(j)} + e_2^{(j)}}{2} \leq u \leq e_2^{(j)} \\
      1 - 2\left(\frac{u-e_3^{(j)}}{e_4^{(j)}-e_3^{(j)}}\right)^2, & e_3^{(j)} \leq u \leq \frac{e_3^{(j)} + e_4^{(j)}}{2} \\
      2\left(\frac{u-e_4^{(j)}}{e_4^{(j)}-e_3^{(j)}}\right)^2, & \frac{e_3^{(j)} + e_4^{(j)}}{2} \leq u \leq e_4^{(j)} \\
      0, & \text{otherwise,} 
   \end{array}
   \right.
\end{equation}
for evaluation of colony area, circularity or expected colony count, $j=1,2,3$ respectively, represented by the variable $u$. Hence, each segmented candidate colony will have its property set $j$ for all points $u \in U$ graded according to the MF \eqref{eq: mu_fuzzy} such that $\mu_j: U\rightarrow [0,1]$. The parameters $e_1^{(j)}$, $e_2^{(j)}$, $e_3^{(j)}$ and $e_4^{(j)}$ are adjustable and correspond to the pi-shaped edges which form the selection space (see \autoref{fig: FuzzyPi}).

For $\mu_1$, the corners of the area distribution are $$\left(e_1^{(1)},e_2^{(1)},e_3^{(1)},e_4^{(1)}\right) = \left(0.5a_{min},a_{min},\max{\left(2a_{min},a_{max}\right)},2a_{max}\right),$$ where $a_{min}$ and $a_{max}$ are minimum and maximum user specified colony sizes, respectively. For $\mu_2$, the circularity parameters are flexible $\left(e_1^{(2)},e_2^{(2)},e_3^{(2)},e_4^{(2)}\right) \left(c_1,c_2,c_3,c_4\right)$, where $0 \leq c_1<c_2<c_3<c_4 \leq 1$ with circularity value 1 for a perfect circle. For the expected count distribution $\mu_3$, the function edges are defined as $\left(e_1^{(3)},e_2^{(3)},e_3^{(3)},e_4^{(3)}\right)=\left(1,E_m,2E_m,3E_m-1\right)$, where $E_m=\left\lceil\frac{a_m}{\widetilde{a}}\right\rceil$, $a_m$ is the area of $BLOB_m$ and $\widetilde{a}$ is the median area of $BLOB_1,\ldots,BLOB_n$. Thus, the multi-feature fuzzy logic presented is utilized to assess the geometrical shapes of subdivided colonies within an iterated $BLOB_m$ after each successive watershed segmentation. This is performed in order to objectively select the segmented outcome that attains colonies of coherent geometrical characteristics. Ultimately, the segmentation procedure yields an appropriate binary image representing the final feature-endorsed colonies, $\mathbf{I}_{seg}$.

\subsection{Experimental setup and data acquisition}

\subsubsection{Parameter selection}

The images are loaded in the ACC algorithm and the parameters are manually tuned as listed in \autoref{tab: parameters} for each dataset. During the PCA acquisition (phase I), the PC images are firstly processed using CLAHE in preparation for the GLCM contrast selection criterion. The contrast enhancement is performed by partitioning each image into $16 \times 16$ regions with a clip limit factor of $0.008$. For the computation of the co-occurrence matrix, $\mathbf{G}$, in equation \eqref{eq: NGLCM} the spatial dependence between neighboring pixels was evaluated at $N_g=64$ gray-levels. Further, the GLCM is highly dependent on the parameters $d$ and $\theta$. Thus, applying equation \eqref{eq: NGLCM}, several matrices was obtained for each change in direction $\theta$. This was defined by four different offset vectors; $[0,d]$ ($\theta=0^\circ$), $[-d,d]$ ($\theta=45^\circ$), $[-d,0]$ ($\theta=90^\circ$), $[-d,-d]$ ($\theta=135^\circ$), where the displacement $d=1$ (in pixels) is set to examine merely adjacent pixels in $\mathbf{J}$ (the PC images). The co-occurrence matrix and thereby the contrast statistic was readily computed for each offset and then averaged. The choice of $d$ is justified as a pixel is more likely to be correlated to closely located pixels than those further away.
%and, thus, capturing more detailed textural information. While $N_g$ is set to extract more accurate textural information, without considerable increase in computational costs.

For the $k$-means acquisition (phase II), the processing stage of $\mathbf{I}_{PCA}$ included morphological opening-closing by reconstruction using a disk-shaped structuring element with a radius of $r_{obrcbr}$ (in pixels), before smoothing using a filter with a 2D Gaussian kernel of size $s_x \times s_y$ (see \autoref{tab: parameters}). These operations were used for background suppression and to smooth the varying spatial image intensity for outliers, respectively. Here, $r_{obrcbr}$ should conform with areas size of the BLOBs as it should be exceedingly greater, whereas $s_x \times s_y$ should reduce evident noise over smaller spatial regions. In the processing step of $\mathbf{I}_{BLOBs}$, various morphological operations were applied on the binary mask such as dilation and flood-filling of holes. 

$\mathbf{I}_{gray}$ was also processed prior to the watershed segmentation. 2D Gaussian filtering (to avoid over-segmentation of the BLOBs) and AHE (to contrast enhance each BLOB) was employed, where the enhanced image was min-max normalized (see \autoref{tab: parameters}). The Gaussian smoothing on $\mathbf{I}_{gray}$ is set to directly affect the forthcoming segmentation of the extracted BLOBs as the filtering is performed on regions in $\mathbf{I}_{gray}$ masked out by $\mathbf{I}_{BLOBs}$. Depending on the image dpi, area size of the actual colonies and colony confluency, the standard deviation of the Gaussian blur of the BLOB grayscale intensities should be chosen accordingly.

During the watershed segmentation (phase III), each masked $BLOB_m$ having an area $a_m > a_{thresh} = 0.6\Tilde{a}$ and circularity $c_m < 0.6$ was further separated through the multi-threshold segmentation. These condition limits for segmentation were kept fixed. Enforcing this, we chose $h_i\in [h_{min},h_{max}]=[0.15,0.37]$ with incremental steps $\Delta_h=0.01$ as a search space for all datasets. The size of this watershed search space has a pronounced influence on the runtime; even though a smaller range and/or larger $\Delta_h$ would yield a shorter computation time, doing so may not ensure optimal segmentation results. Thus, a high colony density necessitates a large search span by lowering the $h_{min}$ value to eventuate a finer segmentation of BLOBs, whilst choosing a very large $h_{max}$ value may not be cost-effective. The pi-shaped MF parameters for the area and circularity distributions were set to $\left(0.5a_{min},a_{min},a_{max},2a_{max}\right)$ and $\left(c_1,c_2,c_3,c_4\right)=\left(0.15,0.5,0.9,1\right)$, respectively, where $a_{min}$ and $a_{max}$ (in pixels) are provided by the user (see \autoref{tab: parameters}). The edges for the expected colony count within each iterated $BLOB_m$, $\left(1,E_m,2E_m,3E_m-1\right)$, are adaptively computed throughout the segmentation process. Subsequent segmented colonies were recursively divided until the criteria $a_m \leq a_{thresh}$ was met.
\begin{table}[ht!]
    \centering
    \begin{tabular}{c c|c c c c}
        \toprule
        \multirow{3}{*}{\makecell{dataset}} & \multirow{3}{*}{specie} & \multicolumn{4}{c}{acquisition parameters} \\
        & & $s_x \times s_y$ & $s_x \times s_y$ & $r_{obrcbr}$ & $(a_{min},a_{max})$ \\
        & & ($\mathbf{I}_{PCA}$) & ($\mathbf{I}_{gray}$) & (pixels) & (pixels) \\
        \midrule
        1 &T-47D & $2\sigma \times 2\sigma$ & $3\sigma \times 3\sigma$ & $40$ & $(40,8000)$ \\
        \midrule
        \multirow{4}{*}{2} & \textit{E. coli} & $3\sigma \times 3\sigma$ &  $10\sigma \times 10\sigma$ & $90$ & $(1000,35000)$ \\
        &\textit{Klebs. pn.} & $3\sigma \times 3\sigma$ &  $6\sigma \times 6\sigma$ & $65$ & $(800,20000)$ \\
        &\textit{Pseud. ae.} & $3\sigma \times 3\sigma$ & $8\sigma \times 8\sigma$ & $80$ & $(2500,20000)$ \\
        &\textit{Staph. au.} & $3\sigma \times 3\sigma$ & $6\sigma \times 6\sigma$ & $30$ & $(500,5000)$ \\
        %\midrule
        %3&A549 & $2\sigma \times 2\sigma$ & $3\sigma \times 3\sigma$ & $30$ & $(140,8000)$ \\
         \bottomrule
    \end{tabular}
    \vskip5pt
    \caption{Parameter selection in the automated colony counting (ACC) method for image segmentation of the different clonogenic species. The Gaussian smoothing filter size, $s_x \times s_y$, specified as a 2-element vector of positive numbers in terms of the standard deviation, $\sigma$, of the Gaussian distribution, is applied on $\mathbf{I}_{PCA}$ and $\mathbf{I}_{gray}$. The radius of the disk-shaped structuring element in the morphological opening-closing by reconstruction, $r_{obrcbr}$, is given in pixels. Minimum and maximum user specified colony areas, $a_{min}$ and $a_{max}$ respectively, are given in pixels.}
    \label{tab: parameters}
\end{table}

\subsubsection{Cell culture and manual counting}

Human breast ductal cell carcinoma cells of the T-47D line were cultured in RPMI medium (Lonza) , supplemented with 10\% FBS (Biochrom), 1\% penicillin/streptomycin (Lonza) and 200 units per liter insulin (Gibco), at 37$^\circ$C in air with 5\% CO\textsubscript{2}. The cells were kept in exponential growth by reculturing twice per week with one additional medium change per week. The seeded number of cells was low which consequently formed sparsely populated colonies in each culture dish. For more information on the cell culture and colony formation assay used in the current work, see e.g. \cite{edin2012low}. 

To validate the quality of the presented ACC segmentation algorithm, we compared the ACC number to the number produced by the recently published method \textit{AutoCellSeg}\cite{torelli2018autocellseg} (both datasets), as well as to the manual colony counting (MCC) facilitated by 3 trained human observers (only dataset 1). Here the observers were independent meaning that no subject could know the results of any other before counting. Additionally, an extra independent observer established a GT by manual counting during a microscopic analysis of the culture dishes for comparison (dataset 1). 

\subsubsection{Data description} \label{data description}

% technical info on images
The ACC algorithm was applied on images of the cell culture flasks containing fixed and stained cell colonies. We conducted experiments on both proprietary and open-source data. 

\begin{figure}[ht!]
    \centering
    \includegraphics[width=0.8\textwidth]{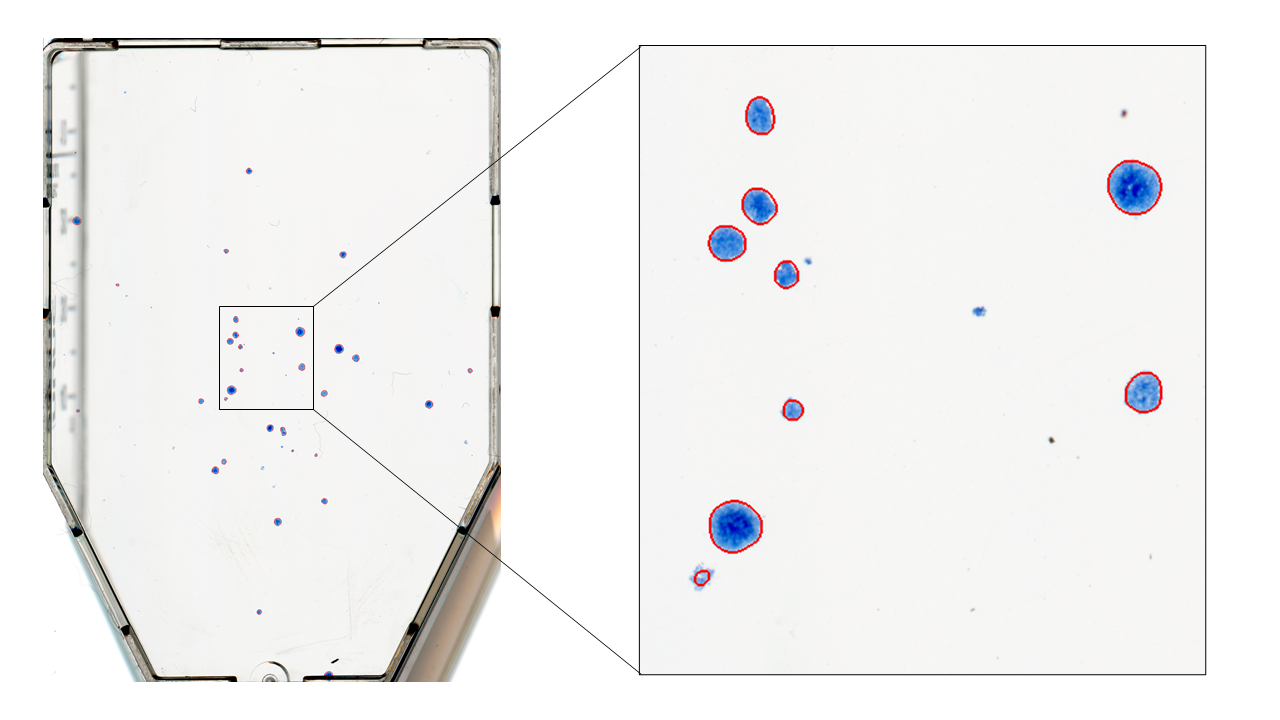}
    \caption{Example image from dataset 1. The segmentation suggested by the automated colony counting (ACC) algorithm is outlined in red.}
    \label{fig:example_img1}
\end{figure}

Proprietary data (dataset 1) were obtained from a flatbed laser scanner (Epson Perfection V850 Pro), providing $rgb$ images with a resolution of 2125$\times$2985, 1200 dots per inch (dpi), 21.17 $\mu$m/pixel spatial resolution and 48-bit depth. No prior filtering nor adjustments were performed on the captured images during scanning with the scanner software (EPSON Scan v3.9.3.3). An example of cell colony image is provided in \autoref{fig:example_img1}. The cell dish contains cell colonies, as well as background structures (e.g. shadows) and outer contours of the cell flask. The segmentation suggested by the ACC is delineated in red. The full dataset consists of 16 cell culture flasks used for a colony formation assay of the T-47D (breast) cancer cell line.

\begin{figure}[!ht]
    \centering
    \begin{subfigure}[c]{0.24\textwidth}
        \centering
        \includegraphics[width=\textwidth]{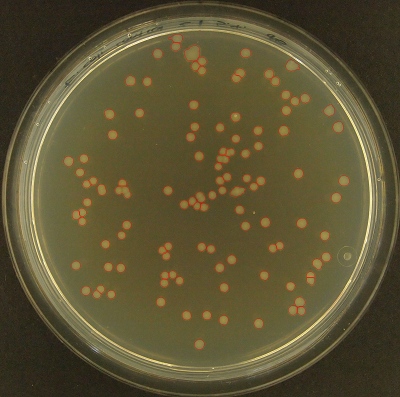}
        \caption{\textit{E. coli}}
    \end{subfigure}
    \begin{subfigure}[c]{0.24\textwidth}
        \centering
        \includegraphics[width=\textwidth]{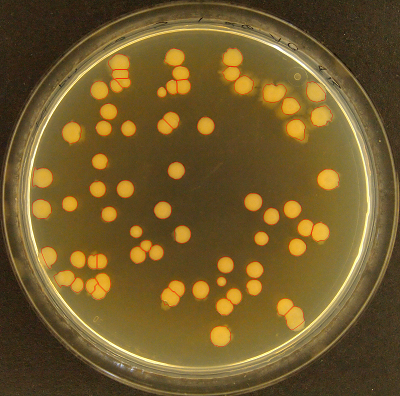}
        \caption{\textit{Klebsiella pneumoniae}}
    \end{subfigure}
    \begin{subfigure}[c]{0.24\textwidth}
        \centering
        \includegraphics[width=\textwidth]{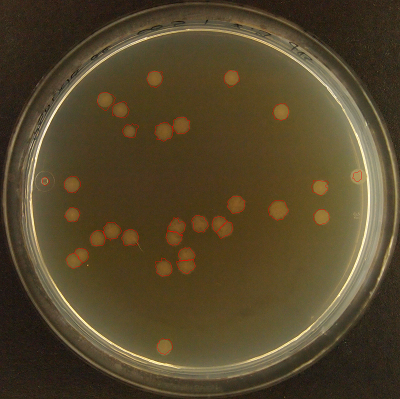}
        \caption{\textit{Pseudomonas aeruginosa}}
    \end{subfigure}
    \begin{subfigure}[c]{0.24\textwidth}
        \centering
        \includegraphics[width=\textwidth]{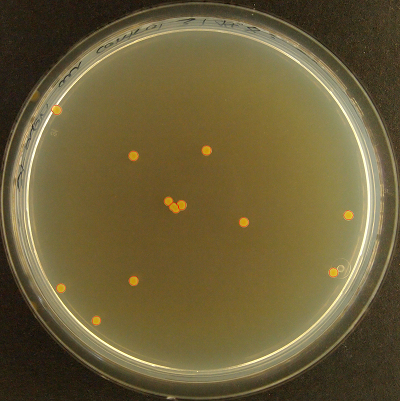}
        \caption{\textit{Staphylococcus aureus}}
    \end{subfigure}
    \caption{Example images from dataset 2. The segmentation suggested by the automated colony counting (ACC) algorithm is outlined in red.}
    \label{fig:example_os_img}
\end{figure}

% write short paragraph about open source data
Open-source images (dataset 2) of $rgb$ color representation, 4032$\times$3024 resolution, 314 dpi, 80.89 $\mu$m/pixel spatial resolution and 24-bit depth, with accompanying GT delineations, were obtained from the publicly available AutoCellSeg's GitHub repository (\url{https://github.com/AngeloTorelli/AutoCellSeg/tree/master/DATA/Benchmark}). The dataset contained 12 images of four bacterial species (3 images each), including \textit{E. coli}, \textit{Klebsiella pneumoniae} (\textit{Klebs. pn.}), \textit{Pseudomonas aeruginosa} (\textit{Pseud. ae.}) and \textit{Staphylococcus aureus} (\textit{Staph. au.}). The GT colony delineations were produced by the authors Torelli et al. using Adobe Photoshop before being converted into binary masks. Delineations obtained for this dataset using the ACC algorithm are shown in \autoref{fig:example_os_img}.

\subsubsection{Hardware}

The segmentation using the ACC procedure was implemented in MATLAB (MathWork, Natick, MA, USA) and executed on an Intel Core i7-8565U CPU @ 1.80 GHz with 16 GB RAM. The average runtime of the proposed algorithm was 114 seconds per image, which is adequate when considering the software as a fully automated batch throughput solution for large datasets. However, runtime optimization and parallelization are not in the scope of this work and will be considered in future projects. The AutoCellSeg results were obtained by installing and utilizing the freely available AutoCellSeg software (\url{https://github.com/AngeloTorelli/AutoCellSeg}), which is based on the open-source implementation by Torelli et al., and run on a partially automated mode via the GUI with similar processing parameters as in our own pipeline.

\subsubsection{Statistical analysis}

In addition to cell colony counts, we investigated the spatial information associated with the detected cell colonies in the images. Hence, \autoref{tab:exp1_results} further provides binary classification metrics for both ACC and AutoCellSeg using a region-wise definition of the confusion matrix. Given the segmentation of ACC or AutoCellSeg, respectively, as well as one centralized coordinate point per colony representing the GT (GT mark), we considered a colony as \textit{detected} if at least one GT mark was within the delineated area. Such regions were denoted as \textit{true positives} ($TP$). We denoted a cell colony as \textit{false positive} ($FP$) if the delineated region did not contain any GT mark. Finally, \textit{false negative} ($FN$) regions were obtained from those GT marks which were either located outside the delineated areas (not detected by the algorithm) or in a delineated region together with other GT marks (merged with other colonies by the algorithm). The $F_1$ score was chosen as a binary classification metric to measure the spatial accuracy of the detected colonies made by the observers and the ACC. Here the $F_1$ score is the harmonic mean between the precision (pre) and recall (rec):
\begin{equation}
    F_1 
    = \frac{2}{\text{pre}^{-1} + \text{rec}^{-1}}
    = \frac{2}{\left(\frac{TP}{TP+FP}\right) ^{-1} + \left(\frac{TP}{TP+FN}\right)^{-1}},
\end{equation}
where the precision measures the ratio of $TP$ cases to the total predicted positive cases ($TP+FP$), while recall measures the ratio of $TP$ cases to the total actual positives cases ($TP+FN$).

\section{Results} \label{sec:experiments}

\subsection{Dataset 1} 

\autoref{tab:exp1_results} shows an overview on the results from ACC, AutoCellSeg and MCC, as well as their respective values compared to the GT on dataset 1. Even though both MCC and GT were obtained from manual counting, the former was based on manual counting on the same images that were presented to the algorithm, whereas the GT is more reliable due to the in-depth information from the microscopy. For each image, the average MCC is shown along with its mean absolute deviation between the observers.
\begin{table}[ht!]
    \centering
    \begin{tabular}{c|c||cccc|cccc|cccc}
        \toprule
        \multirow{2}{*}{image} & GT & \multicolumn{4}{c |}{ACC} & \multicolumn{4}{c |}{AutoCellSeg} & MCC \\
        & cnt &  pre & rec & $F_1$ & cnt & pre & rec & $F_1$ & cnt & cnt \\
        \midrule
        1  &37& 0.97 & 0.89 & \textbf{0.93} &34& 0.87 & 0.70 & 0.78 & 30 & 36.7 $\pm$ 2.4 \\
        2  &48& 1.00 & 0.88 & \textbf{0.93} &42& 0.89 & 0.83 & 0.86 & 45 & 48.0 $\pm$ 1.3 \\
        3  &45& 0.98 & 0.87 & \textbf{0.92} &40& 0.87 & 0.87 & 0.87 & 45 & 45.3 $\pm$ 1.8 \\
        4  &61& 0.95 & 0.90 & 0.92 &58& 0.93 & 0.93 & \textbf{0.93} & 61 & 62.0 $\pm$ 2.7 \\
        5  &54& 0.96 & 0.87 & \textbf{0.91} &49& 0.80 & 0.83 & 0.82 & 56 & 51.0 $\pm$ 2.0 \\
        6  &49& 0.98 & 0.80 & \textbf{0.88} &40& 0.85 & 0.80 & 0.82 & 46 & 47.3 $\pm$ 2.2 \\
        7  &36& 0.91 & 0.81 & \textbf{0.85} &32& 0.76 & 0.72 & 0.74 & 34 & 35.7 $\pm$ 2.9 \\
        8  &33& 0.97 & 0.88 & \textbf{0.92} &30& 0.76 & 0.76 & 0.76 & 33 & 30.3 $\pm$ 1.8 \\
        9  &45& 0.95 & 0.80 & \textbf{0.87} &38& 0.82 & 0.62 & 0.71 & 34 & 40.7 $\pm$ 1.8 \\
        10 &64& 0.94 & 0.73 & \textbf{0.82} &50& 0.93 & 0.61 & 0.74 & 42 & 59.3 $\pm$ 3.6 \\
        11 &34& 0.97 & 0.94 & \textbf{0.96} &33& 0.86 & 0.91 & 0.89 & 36 & 33.3 $\pm$ 1.1 \\
        12 &40& 0.97 & 0.90 & \textbf{0.94} &37& 0.86 & 0.75 & 0.80 & 35 & 40.7 $\pm$ 2.2 \\
        13 &40& 0.93 & 0.68 & \textbf{0.78} &29& 0.86 & 0.63 & 0.72 & 29 & 37.0 $\pm$ 2.7 \\
        14 &52& 0.94 & 0.92 & \textbf{0.93} &51& 0.91 & 0.81 & 0.86 & 46 & 52.3 $\pm$ 1.8 \\
        15 &52& 0.96 & 0.92 & \textbf{0.94} &50& 0.85 & 0.77 & 0.81 & 47 & 54.3 $\pm$ 1.8 \\
        16 &48& 1.00 & 0.81 & \textbf{0.90} &39& 0.88 & 0.75 & 0.81 & 41 & 46.0 $\pm$ 2.0 \\
        \bottomrule
    \end{tabular}
    \vskip5pt
    \caption{Results for T-47D cell dishes (dataset 1), obtained from automated colony counting (ACC) via the presented procedure, the \textit{AutoCellSeg} method, as well as manual colony counting (MCC), compared to the ground truth (GT). Manual counts are averaged over 3 independent observers $\pm$ mean average deviation. Estimates for precision (pre), recall (rec), $F_1$ score and colony count (cnt) produced by each method for each image in dataset 1 are compared.}
    \label{tab:exp1_results}
\end{table}

The counts obtained from all methods achieve similar results and do not show a clear winner: our proposed ACC method produced a root-mean-square error (RMSE) of 14\% with a tendency to underestimate the GT count. AutoCellSeg showed similar characteristics with a RMSE of 17\%. Although the MCC had a similar RMSE (ACC errors are within the error bounds associated with MCC), the manual observers slightly overestimated the colony number: in all except for three images (images 11, 12, 13), the mean MCC was higher than the GT count.

With regard to spatial information, ACC obtained superior $F_1$ scores than AutoCellSeg, although the absolute ranges for both procedures were on a very high level ($F_1$ score mostly $>90$\%). This indicates that ACC can outperform the current state-of-the method. Analyzing the metrics in detail revealed that in most cases, both precision and recall could be improved by ACC (e.g., image 10 and 13). In few cases, we observe that ACC obtains a higher $F_1$ score, although the error with respect to absolute colony counts is higher compared to AutoCellSeg (e.g. image 2). This anomaly might be caused by a mutual compensation of different error types in AutoCellSeg, such as dividing one cell colony into multiple regions and neglecting others at the same time. This will decrease the $F_1$ score, but remain undisclosed when comparing overall colony counts.

\subsection{Dataset 2}

In addition to the results obtained from the proprietary T-47D cell dataset, we used both algorithms, ACC and AutoCellSeg, on publicly available open-source datasets. The datasets differ from dataset 1 in coloring, shape of the cell dish, size of the investigated cell colonies, image resolution and background. Evaluation is made in the same way as for dataset 1, except for that no manual counting from different observers were available for evaluation. The results are shown in \autoref{tab:exp2_results}.
\begin{table}[ht!]
    \centering
    \begin{tabular}{c|c || cccc | cccc}
        \toprule
        \multirow{2}{*}{\makecell{specie}} & GT & \multicolumn{4}{c |}{ACC} & \multicolumn{4}{c}{AutoCellSeg}\\
        & cnt & pre & rec & $F_1$ & cnt & pre & rec & $F_1$ & cnt \\ 
        \midrule
        \multirow{3}{*}{\makecell{\textit{E. coli}} } & 116 & 0.99 & 0.96 & \textbf{0.97} & 112 & 0.98 & 0.97 & \textbf{0.97} & 114 \\ 
        & 80 & 0.97 & 0.94 & \textbf{0.96} & 77 & 0.9 & 0.96 & 0.93 & 86 \\ 
        & 32 & 0.94 & 1.00 & \textbf{0.97} & 34 & 0.8 & 1.00 & 0.89 & 40 \\
        \midrule
        \multirow{3}{*}{\makecell{\textit{Klebs. pn.}} } & 67 & 0.99 & 0.99 &  \textbf{0.99} & 67 & 1.00 & 0.97 & 0.98 & 64 \\ 
        & 49 & 1.00 & 0.94 &  \textbf{0.97} & 46 & 0.94 & 0.9 & 0.92 & 47 \\ 
        & 27 & 0.96 & 1.00 & \textbf{0.98} & 28 & 0.96 & 1.00 & \textbf{0.98} & 28 \\ 
        \midrule
        \multirow{3}{*}{\makecell{\textit{Pseud. ae.}}} & 29 & 1.00 & 1.00 &  \textbf{1.00} & 29 & 0.97 & 0.97 & 0.97 & 29 \\ 
        & 20 & 1.00 & 1.00 & \textbf{1.00} & 20 & 0.95 & 0.95 & 0.95 & 20 \\ 
        & 25 & 0.96 & 0.92 & 0.94 & 24 & 1.00 & 0.96 & \textbf{0.98} & 24 \\ 
        \midrule
        \multirow{3}{*}{\makecell{\textit{Staph. au.}}}  & 13 & 1.00 & 0.92 & \textbf{0.96} & 12 & 1.00 & 0.85 & 0.92 & 11 \\ 
        & 106 & 0.97 & 0.94 & \textbf{0.96} & 103 & 0.95 & 0.88 & 0.91 & 98 \\ 
        & 88 & 0.99 & 0.95 & \textbf{0.97} & 85 & 0.98 & 0.93 & 0.95 & 84 \\ 
        \bottomrule
        \end{tabular}
        \vskip5pt
    \caption{Results for open-source cell dish images (dataset 2), obtained from automated colony count (ACC) via the presented procedure, as well as the \textit{AutoCellSeg} method, compared to the ground truth (GT). Estimates for precision (pre), recall (rec), $F_1$ score and colony count (cnt) produced by each method for each image in dataset 2 are compared.}
    \label{tab:exp2_results}
\end{table}

The experiment conducted on dataset 2 demonstrates that ACC is able to outperform AutoCellSeg in 9 out of 12 cases with respect to $F_1$ scores and performs equally well in 2 case (image control 2 of \textit{E. coli} and image control 3 of \textit{Klebsiella pneumoniae}), whereas AutoCellSeg scored higher on only one case (image 2 of \textit{Pseudomonas Aeruginosa}). Indirectly, the presented results can be compared to experiments from \cite{torelli2018autocellseg} on the same datasets, where other state-of-the-art methods are evaluated. Unlike for dataset 1, the single images in this experiment show more variability, hence the high-quality results underline the flexibility of the presented algorithm.

\section{Discussion}

A clear benefit of the proposed algorithm is the saving of resources in terms of time and manual effort. Remarkably, the algorithm matches manual observation techniques not only in terms of speed, but also delivers robust and objective results.

Our experiments demonstrated that the proposed algorithm is capable of solving the automated cell counting problem and serves as a valid alternative to manual procedures with a competitive quality. Herein, the PC image containing the color variability of the colonies offers a reliable and selective depiction of the colonies when compared to the traditional grayscale image, $\mathbf{I}_{gray}$, of $\mathbf{I}$. Without PCA, feature extraction from $\mathbf{I}_{gray}$ is liable to include and segment falsely detected objects with similar grayscale intensities as colonies. Also, the results are superior to those obtained from the AutoCellSeg state-of-the-art method and in the range of human inter-observer variance. Thus, further refinement is hardly possible unless more accurate reference data are available. Particularly the flexibility of our presented ACC algorithm, taking different cell dish geometries, background, image resolution and coloring into account, proved its high value.

We discovered a small bias between the human observers and the automated counts, particularly on dataset 1. In this case, the algorithm tends to provide lower estimates. A manual evaluation showed that particularly small and sparsely populated cell regions with low contrast to the background were neglected by the automated algorithm in specific cases, but identified as colonies by human observers. Such errors can be reduced by parameter tuning, particularly those related to watershed segmentation. However, the fact that the results from different human observers are not always consistent (in particular when judging such small regions) shows the challenges of the task. Following the definition of cell colonies as conglomerations of more than typically 50 cells, this threshold can solely be verified by microscopy. Enhanced parameter tuning procedures to fit different problem setups will be investigated in future work when reliable GT information is available for a larger amount of data.

In addition, identification of the centroid coordinates of each colony listed together with information about respective colony ID, area, circularity and mean/standard deviation of intensity (color, grayscale and PCs) distribution as well as colony count are saved for further analysis upon completion of our segmentation procedure. Moreover, a binary mask containing fully filled areas representing the segmented colonies is also saved for each image. Thus, the culminated output from the algorithm could open for new applications with colony formation assays beyond regular colony counting. This is useful for users who, for instance, wish to evaluate the colony size of a distinct cell population with respect to treatment efficacy of e.g. a drug or irradiation dose conducted in a microbiological or radiobiological experiment.

Compared to other contemporary problems in digital image processing and computer vision, the available amount of training and test data is very limited and the GT is not completely unbiased. Hence, complex models such as DCNNs are hardly applicable. Instead, the presented algorithm is unsupervised and overcomes the limitations imposed from the training data by building on well-established and easy-to-train components. An extension with other architectures will be evaluated when more training data are available in the future. Translating the proposed algorithm into other languages such as Python, R etc. is also valuable as it allows for more flexibility to extend the program in various programming languages with their complementary packages or modules.

\section{Conclusion}
We presented a novel algorithm to segment cell colonies on images of cell dishes from colony formation experiments. Our ACC procedure is based upon a tailored pipeline with three major components: PCA bundles the information content from the $rgb$ color channels, $k$-means clustering identifies conglomerate areas of cell colonies and a fuzzy statistics modification of the watershed algorithm splits them into separate cell colonies.

Our experiments were conducted on a breast cancer cell line as well as publicly available images from other cell types. In our analyses, the method was evaluated against both a recent state-of-the-art method and manual counting by human experts. The experiments demonstrated that the proposed algorithm is able to beat the benchmark, as well as it meets the expectations by obtaining results of similar quality as the manual observers.

\section*{Acknowledgments}
We would like to thank Julia Marzioch, Olga Zlygosteva and Magnus Børsting from the Department of Physics at the University of Oslo for conducting the manual counting of the cell colonies.
This work was supported by the South-Eastern Norway Regional Health Authority (Project ID 2019050) and the Norwegian Cancer Society (Grant ID 182672).

\bibliographystyle{unsrt}  
\bibliography{references}

\end{document}